\documentclass{article}
\usepackage{ijcai23}

\usepackage{amsmath} 
\usepackage{amssymb} 
\usepackage{amsfonts} 
\usepackage{multirow} 
\usepackage{color}

\usepackage{times}
\usepackage{soul}
\usepackage{url}
\usepackage[hidelinks]{hyperref}
\usepackage[utf8]{inputenc}
\usepackage[small]{caption}
\usepackage{graphicx}
\usepackage{amsmath}
\usepackage{amsthm}
\usepackage{booktabs}
\usepackage{algorithm}
\usepackage{algorithmic}
\usepackage[switch]{lineno}
\usepackage{color}

\title{STS-GAN: Can We Synthesize Solid Texture with High Fidelity \\from Arbitrary 2D Exemplar?}

\author{
Xin Zhao$^{1}$\and
Jifeng Guo$^4$\and
Lin Wang$^{2,3,}$\thanks{Corresponding authors: Lin Wang (wangplanet@gmail.com), Bo Yang (yangbo@ujn.edu.cn)}\and
Fanqi Li$^{2,6}$\and
Jiahao Li$^2$\and
Junteng Zheng$^5$\and
Bo Yang$^{3,2,*}$
\affiliations
$^1$ Shandong Provincial Key Laboratory of Preparation and Measurement of Building Materials,\\ University of Jinan, Jinan 250022, China\\
$^2$ Shandong Provincial Key Laboratory of Network Based Intelligent Computing,\\ 
University of Jinan, Jinan 250022, China\\
$^3$ Quan Cheng Laboratory, Jinan 250100, China\\
$^4$ School of Computer Science and Engineering, South China University of Technology,\\ Guangzhou 510641, China\\
$^5$ David R. Cheriton School of Computer Science, University of Waterloo, \\ Waterloo, ON N2L 3G1, Canada\\
$^6$ Shandong Qiuqi Analysis Instrument Co., Ltd., Jinan 250024, China
}
\begin{document}

\maketitle
% \color{blue}
\begin{abstract}
    Solid texture synthesis (STS), an effective way to extend a 2D exemplar to a 3D solid volume, exhibits advantages in computational photography. However, existing methods generally fail to accurately learn arbitrary textures, which may result in the failure to synthesize solid textures with high fidelity. 
    In this paper, we propose a novel generative adversarial nets-based framework (STS-GAN) to extend the given 2D exemplar to arbitrary 3D solid textures. 
    In STS-GAN, multi-scale 2D texture discriminators evaluate the similarity between the given 2D exemplar and slices from the generated 3D texture, promoting the 3D texture generator synthesizing realistic solid textures.
    Finally, experiments demonstrate that the proposed method can generate high-fidelity solid textures with similar visual characteristics to the 2D exemplar.
\end{abstract}

\section{Introduction}
\emph{Solid texture synthesis}, a technique for mapping 2D textural information to 3D solid, has been applied widely in computer graphics and vision. In particular, the synthesis process relies only on a few 2D exemplars, generally one 2D exemplar.

\emph{Solid texture synthesis}, in general, extends the visual characteristics of a 2D exemplar into an object whose voxels belong to a volumetric domain $\mathcal{D} \subset \mathbb{R}^3$, sharing a similar internal appearance with the 2D exemplar. During the synthesis process, the color of each voxel in the generated solids is gradually modified by matching exemplars, describing specific appearance properties. 
Eventually, the overall appearance of the synthesized solid is expected to be similar to the given 2D texture exemplar.

During the last several decades, solid texture synthesis attracted a lot of attention in 3D visualization  and volume rendering \cite{Visualizationmodeling,renderapp1,renderapp2}.  It has also received numerous successful stories in real-world applications, such as material science \cite{microsolidapp2}, medical analysis \cite{medicalapp1}, and game development \cite{gamesolidapp1}.

\subsection{Motivation}
Textures usually refer to the visual or tactile experience composed of repeating similar patterns, formally defined as locally stationary, ergodic, stochastic processes \cite{Wei,Texturedefine}. The textures in the real world typically have three characteristics: \emph{Local Markov property}, \emph{Multiscality}, and \emph{Diversity}. (1) \emph{Local Markov property} means the spatial coherence is highly localized  in the neighbourhood. (2) \emph{Multiscality} suggests the spatial coherence could exist at different scales in different ways.  (3) \emph{Diversity} means the style domain of patterns in the real world could be vast. 
Thus, an STS method needs to map the appearance of a 2D exemplar into a 3D solid texture satisfying the three characteristics simultaneously.

The traditional STS methods, like statistical feature matching methods \cite{Heeger1995,GD95,Jagnow2004stereological} or Markov random field-based methods \cite{Wei,Kopf2007,Chen2010},  have received credits in many fields \cite{07kopfapp,solidapp3}. 
Despite some successful stories, these methods fail in accurately projecting 2D appearance into a 3D solid on account of the low expressive power of the model and complexity of real-world applications.

In 2020, \cite{20CNN} introduced neural networks into solid texture synthesis, which may herald a fruitful direction. They proposed a convolutional neural network (CNN)-based method to synthesize solid textures, taking full advantage of its hierarchical expressive capability and extracting features using the VGG \cite{VGG} feature maps. The visual effects of synthesized volumes are at least comparable to the state-of-the-art methods. Similarly, using VGG statistical features, \cite{henzler2020cnn} also provided another point operation-based neural network solution, which can efficiently synthesize 3D textures. 

These neural network-based STS methods are trained to match features, such as VGG statistics. However, the \emph{diversity} of textures makes it challenging to fit different appearances with fixed features \emph{a priori}. It is almost impossible to capture an infinite number of textural appearances with a limited number of features.

The Generative Adversarial Nets (GANs) \cite{NIPSGAN} have been proven to be an effective universal distribution learner, generating diverse images \cite{PSGAN,SINGAN}. 
Following \emph{point operation} strategy from  \cite{henzler2020cnn}, the GramGAN \cite{GramGAN} enables synthesizing solid textures without matching fixed features with generative adversarial nets.
Despite its adaptability to diverse textures, the adopted point operation in GramGAN, simply providing spatial information as network inputs, often leads to difficulty learning complex spatial coherence. It is hard to capture the \emph{local Markov property} and \emph{multiscality} of textures, failing in generating structured textures or complicated stochastic structures \cite{GramGAN}.

\textbf{Question}  \emph{can we synthesize high-fidelity solid texture, capturing all three characteristics, to faithfully reflect the true 3D appearance of arbitrary 2D exemplars?}

Yes, we can. Aiming to address this issue, we propose a novel GAN-based framework for solid textures synthesis, STS-GAN.

\subsection{Contribution}  
We summarize the main contributions of this paper:

\begin{itemize}
 \item STS-GAN extends CNN-based STS to enable learning arbitrary texture distribution and to adapt to textural \emph{diversity}.
 \item Aiming at the \emph{multiscality} of solid texture, a multi-scale strategy is adopted to encourage learning textures hierarchically.
 \item Experiments exhibit our method generates solid textures with higher fidelity than the other state-of-the-art ones.
\end{itemize}

\begin{figure*}[ht]
  \centering
  \includegraphics[width=1\textwidth]{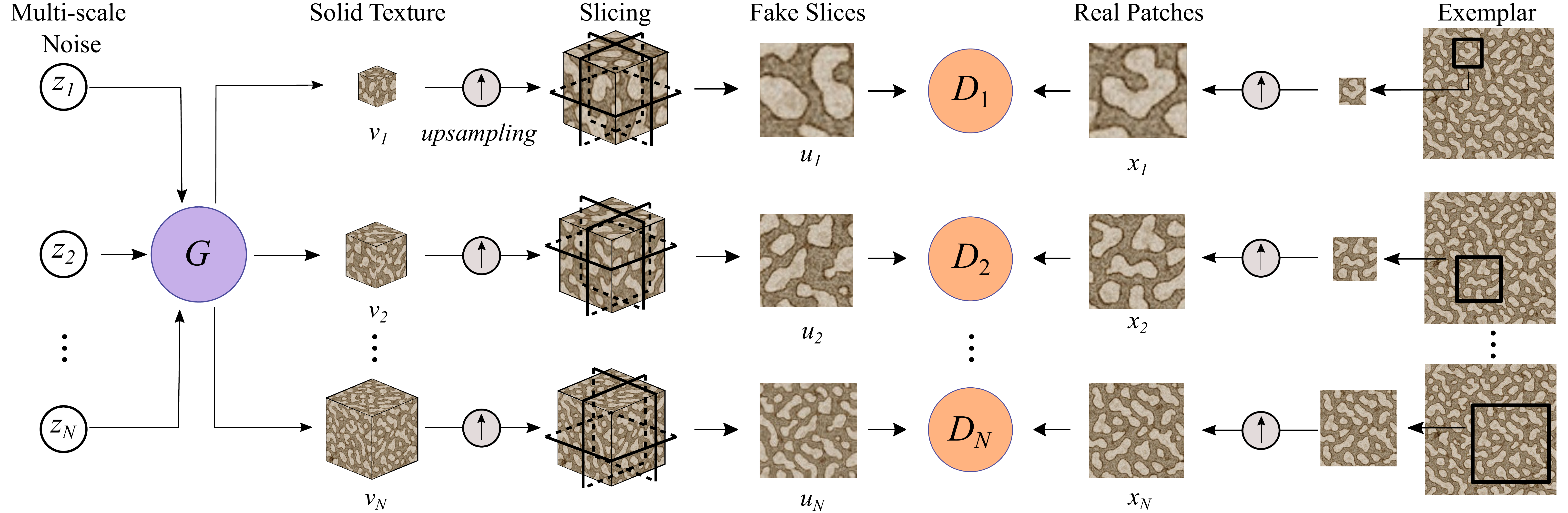}
  \caption{The framework of STS-GAN.
  It is a hierarchical framework containing $N$ learning scales.
  The 3D \emph{solid texture generator} $G$ synthesizes solid textures \{$v_1$, \dots, $v_N$\} by processing multi-scale noises \{$z_1$, \dots, $z_N$\}. 
  At each learning scale $n$, synthesized solid textures $v_n$ and patches $x_n$ which randomly cropped from a given exemplar are upsampled to a same resolution.
  The fake slices $u_n$ in the synthetic solid  $v_n$ are selected at random from the orthogonal directions.
  Finally, the 2D \emph{slice texture discriminators} $D_n$ distinguishes between the slices $u_n$ and the real patches $x_n$. }
  \label{total-structure}
\end{figure*}

\section{Related Works}
\label{related work}
Solid texture synthesis has attracted considerable research interest in the field of computer graphics and vision since it was proposed by \cite{perlin1985} and  \cite{peachey1985solid}. 
In this field, the procedural methods were the earliest family with the advantage of low computational cost. They synthesize textures using a function of pixel coordinates and a set of manually tuning parameters. 
{As perhaps the most famous example, the Perlin Noise \cite{perlin1985} is a smooth gradient noise function that is used to create pseudo-random patterns by perturbing mathematical equations.}

Nevertheless, determining a suitable set of parameters for the desired texture necessitates tedious trial-and-error. 
Furthermore, the semantic gap inhibits people from linking notions such as marble or gravel with accurate parameters.

By contrast, exemplar-based STS methods can generate a new 3D texture from a given 2D exemplar without relying on the artificially accurate texture description. The following is a brief review of various families of these methods.

\subsection{Statistical Feature-Matching Methods}  
These methods use a set of statistical features extracted from a given texture and apply it to solid textures.
The pyramid matching \cite{Heeger1995} method pioneered the work on solid texture synthesis from 2D exemplars, using an image pyramid to capture the characteristics of textures at various resolutions. It is useful to create stochastic textures.
\cite{GD95} presented a solid texture generation method based on the spectral analysis of a 2D texture in various types. \cite{Jagnow2004stereological} proposed a solid texture synthesis method using stereoscopic techniques, which effectively preserve the structure of texture. 

Textures, in general, are diverse and complicated. Statistical feature-based methods tend to synthesize specific textures based on the certain image feature but fail to work on a broad set of textures.

\subsection{Markov Random Field-based Methods}   
These methods model texture as a Markov Random Field (MRF), where each pixel in a texture image depends only on the pixels in its surrounding neighborhood.
Based on the non-parametric MRF model \cite{efros1999texture}, Wei et al.  applied the nearest neighborhood matching strategy coupled with an image pyramid technique to synthesize solid textures \cite{Wei,WeiTSVQ}.

\cite{Kopf2007} synthesized 3D solid textures by adopting MRF as a similarity metric. In this method, the color histogram matching forces the global color statistics of the synthesized solid to match those of exemplars. \cite{Chen2010} integrated position and index histogram matching into the MRF optimization framework using the k-coherence search \cite{k-coherent}, effectively improving the quality of synthetic solids. In general, while these MRF-based techniques may capture hierarchical texture features and generate outstanding results, the conflict between texture diversity and the difficulty of learning a non-parametric MRF model precludes them from producing high-quality solid textures.

\subsection{Neural Nets-based Methods}  
Recently, neural networks have been used to synthesize solid textures because of their capability to approximate any nonlinear functions.

To synthesize realistic volumetric textures, a CNN-based method \cite{20CNN} has been introduced, taking advantage of CNN's powerful expressive capability for spatial autocorrelation data. It takes part of VGG-19 as an image descriptor to conceptualize features extracted from an exemplar. The results prove that it can generate a solid texture of arbitrary size while reconstructing the conceptualized visual features of an exemplar along with some directions.  
In 2020, \cite{henzler2020cnn} also provided another point operation-based neural network solution, which is a generative model of natural textures. The model feeds multiple transformed random 2D or 3D fields into a multi-layer perceptron that can be sampled over infinite domains.

Nevertheless, the \emph{diversity} of textures makes it \emph{hard to use a limited number of VGG features to fit an infinite number of appearances}. Thus, these methods may not accurately capture and extend arbitrary exemplars' texture properties.

As perhaps the pioneer of the GAN-based STS method, GramGAN combined ideas from style transfer and generative adversarial nets to generate realistic 3D textures.
Following the idea of point operation strategy, it takes spatial position information as the input to the generative model.

\subsection{Challenge}
Statistical feature-matching methods rely on features, introducing strong prior and limiting their diversity of applicable textures. 
Although MRF-based methods, taking advantage of their \emph{local Markov property}, potentially can generate diversified 3D textures, their inferior learning capability makes it difficult to accurately estimate the conditional probabilities for the 3D neighborhood from a 2D exemplar.

In terms of neural net-based methods, they provide impressive efficacy due to their powerful expressive power.
Nevertheless, the CNN-based method and work of Henzler et al. cannot always be applicable to \emph{diverse textures}, as they also try to match features such as VGG statistics.
Although GramGAN exhibits its adaptability to diverse textures, the adopted point operation strategy hard to capture \emph{local Markov
property} and \emph{multiscality}, as it simply provides spatial information as network inputs. 
Thus, GramGAN fails to generate structured textures or complicated stochastic structures\cite{GramGAN}.

Therefore, an STS-method, which can synthesize high-fidelity solid textures from arbitrary exemplars, satisfying \emph{local Markov property}, \emph{multiscality}, and \emph{diversity}, is highly desired.

\section{Methodology}
\label{method}
\subsection{Cross Dimensional Appearance Association}

In the beginning, we need to answer the most fundamental question of solid texture synthesizer: \emph{how to associate the appearance of a 3D solid with a given 2D exemplar?}

The solid textural appearance can be described by a joint distribution in the volumetric domain $\mathcal{D}$. This joint distribution can be further decomposed of distributions along distinct directions. Each of these distributions describes the specific appearance in that direction.

Given that we are interested in the texture along a certain direction, the appearance of cross-sections belonging to this direction is drawn from the same distribution, describing the common textural properties. 
As a result, we can learn a joint distribution, in which subdistribution along a corresponding direction reflects the appearance of the given 2D exemplar.  
If we can collect representative exemplar from different directions, we can thus map the textural appearance in the 2D exemplars into 3D domain $\mathcal{D}$ by learning this joint distribution.
Particularly, for an isotropic solid texture, subdistributions of all directions should be the same. 
In addition, it should be noted that the textural appearance usually exhibits different properties at different scales, implying the difference of distributions between scales.

In order to learn the joint distribution, we need to learn the subdistribution along different directions. 
Thus, a \emph{slicing} strategy is adopted to associate 3D solid with the given 2D texture, playing the role of the cross-dimension junction. 
In this strategy, each candidate synthesized solid texture is randomly sliced along different directions to obtain  "fake" cross-sections, which can be used to compare their appearance with a given exemplar directly in the desired directions.
Intuitively, if a randomly sliced cross-section shares the same appearance with the 2D exemplar, the synthesized solid texture and the corresponding "real" solid texture of the 2D exemplar draw from the same joint distribution.

Normally, for anisotropic solid textures, the slicing strategy is operated along given directions. However, if the target solid texture is isotropic, it is sliced orthogonally, as the spatial autocorrelation in 3D space may result in the redundancy of information between non-orthogonal directions.

\subsection{STS-GAN Framework}
\label{subsection:Framework}
To improve the \emph{diversity} of applicable textures,  this work designs a GAN-based framework for synthesizing arbitrary 3D textural appearances by learning texture distribution in the given 2D exemplar. 
The framework of STS-GAN is described in Figure \ref{total-structure}, consisting of 3D \emph{solid texture generator} (STG), 2D \emph{slice texture discriminators} (STDs), and the slicing strategy as junction.

We first define a synthesizing resolutions set $S = \{S_1, S_2, ..., S_N\}$, where $|S|=N$, with the intention of learning textural information at various scales. 
Here, $N$ represents the number of learning scales, $n$ represents the $n^{th}$ scale ($n \in \left[ 1, N \right] $), and the resolution at scale $n$ is $S_n$. 
The concept of scale is generally related to the hierarchical levels of detail. 
Since an image may exhibit different appearances at different scales, we use scaling operation to unify the  resolution to observe a texture at different scales.

As a universal distribution learner, GAN is adopted to learn the distribution of solid texture at 3D domain $\mathcal{D}$. 
In the framework of STS-GAN, the STG plays the role of 3D textures synthesizer, while STDs play the role of critic for discriminating the fidelity of 2D cross-sections in 3D textures. 
The objective of STG $G$  is to synthesize a ``fake'' solid whose slice $u_n$ is similar to the real patch $x_n$, which is randomly cropped from the given exemplar at the corresponding scale $n$. It processes a group of input noises $z_n$ to synthesize a solid texture $v_n$ at scale $n$, 
\begin{equation}
v_n = G(z_n),
\end{equation}
Meanwhile, as STG's opponents, the STDs consist of a collection of multi-scale discriminators $\{D_1, \dots, D_N\}$. $D_n$ learns to differentiate randomly sliced cross-section ${u}_n$ of solid texture $v_n$, from the real patch $x_n$ at scale $n$. 
To observe the appearance of texture at different scales, we upsample those patches with different resolutions to the same resolution.
Moreover, the synthesized 3D solids are upsampled to the same resolution as the corresponding patches to ensure learning texture distribution at the same scale.

In general, the STG generates realistic 3D textures whose sections appear indistinguishable from the given exemplar. By contrast, each STD attempts to distinguish cross-sections of generated 3D texture (fake sample) from the 2D exemplar (the real one) at the corresponding scale. 
{For STG and STDs, we will describe them in the later sections, respectively.}

\begin{figure}[t]
 \centering
		\includegraphics[width=0.98\columnwidth]{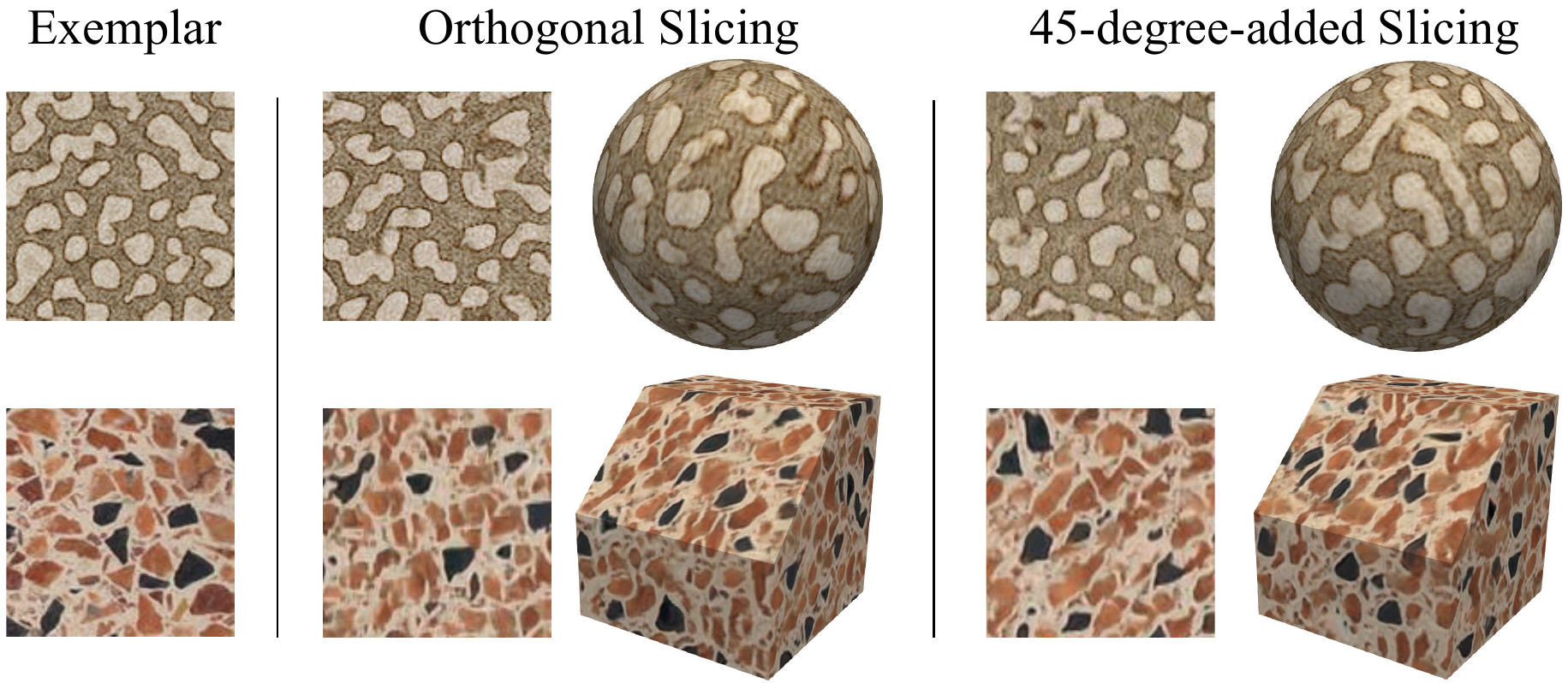}
		\caption{The model's performance using different slicing strategies in training. The middle section presents the outcome of the orthogonal slicing strategy, and the results with the 45 degree-added slicing strategy are shown on the right. In particular, we offer the carved solid and the 2D slice at 45 degree selected randomly from the volume respectively.}
		\label{45du}
\end{figure}

\begin{figure}[t]
 \centering
		\includegraphics[width=0.98\columnwidth]{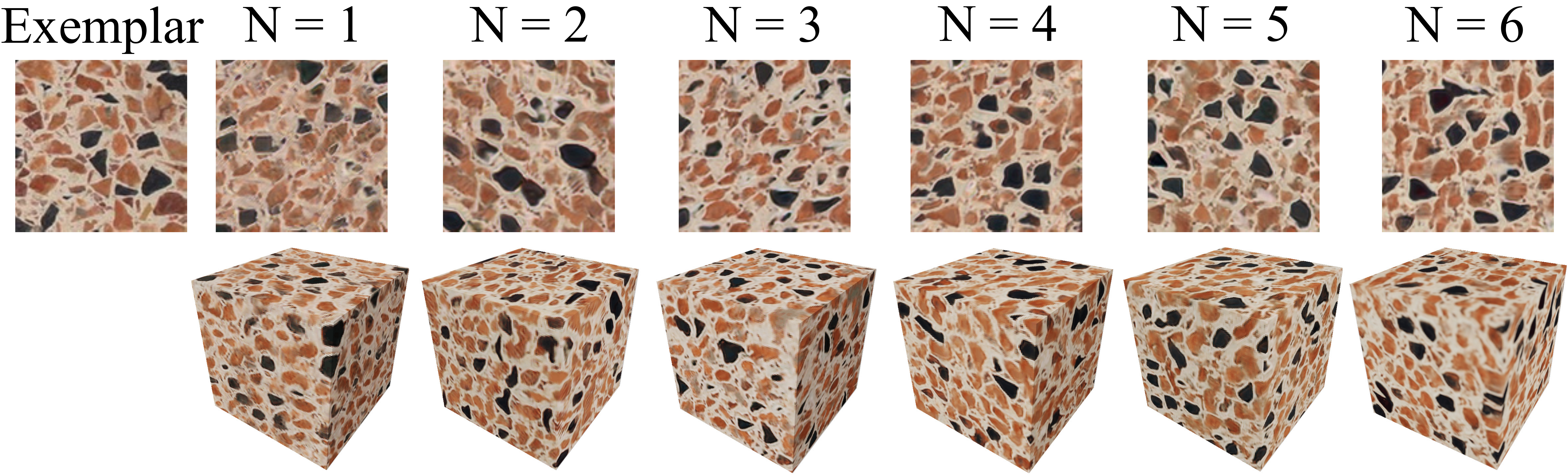}
		\caption{The model's performance with different number of learning scales. The generated solids of each model are shown on bottom, with randomly picked slices from them on the top.}
		\label{amd-1}
\end{figure}

\begin{figure*}[t]
	\centering
		\includegraphics[width=0.95\textwidth]{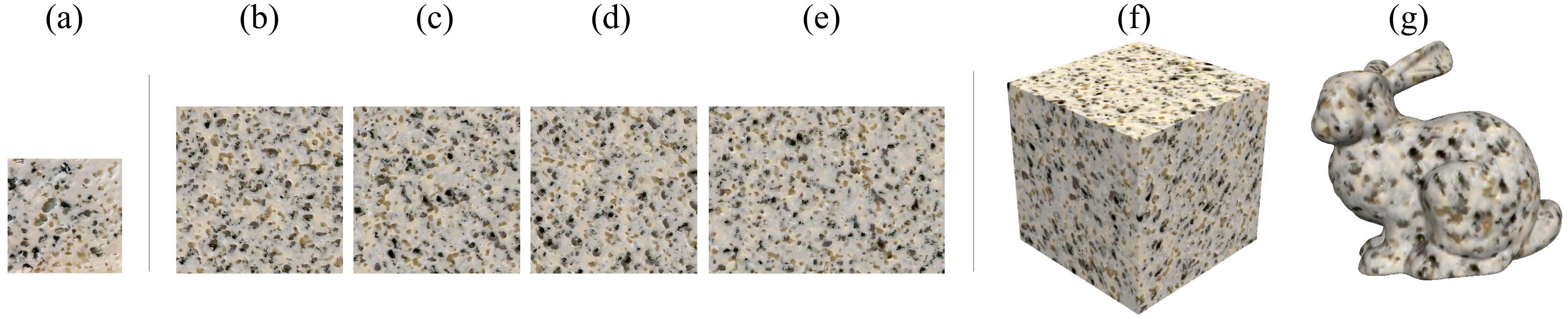}
		\caption{The performance of STS-GAN on isotropic exemplars. (a) texture exemplar, (b) - (d) the slices of the generated solid across the three orthogonal directions, (e) the 45-degree slice, (f) the synthetic solid texture, and (g) texture mapping.  Source: the 3D mesh models are from Stanford 3D Scanning Repository. The resolution of the exemplar is 256, and the resolution of the generated slices and solid in the experiment is 1.5 times that of the original exemplar (i.e., 384).
        }
		\label{isotropic}
\end{figure*}

\begin{figure}[t]
	\centering
\includegraphics[width=0.95\columnwidth]{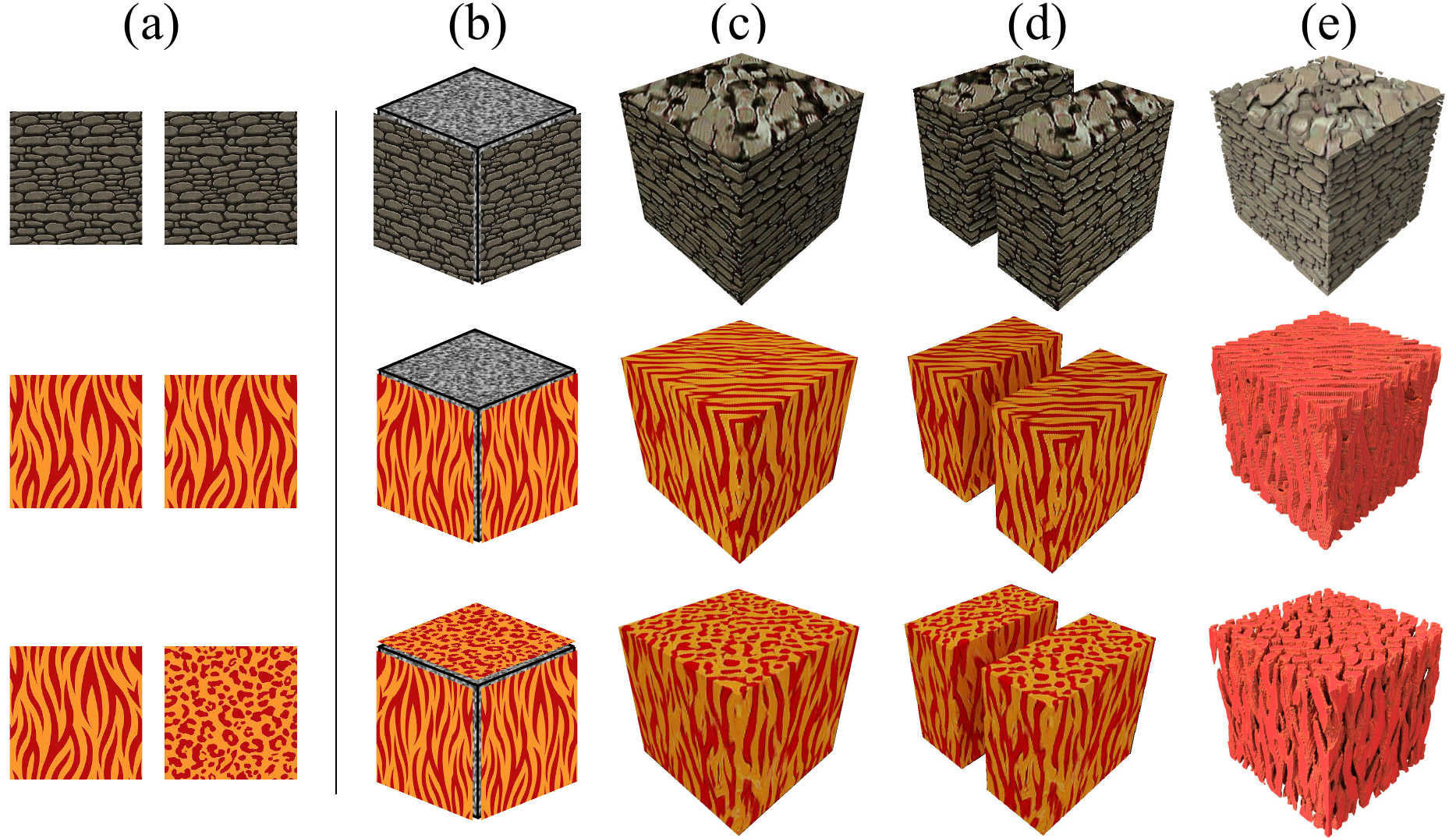}
		\caption{Results of the STS-GAN on different anisotropic exemplars.
		(a) exemplars, (b) learning configurations, (c) synthetic solids,
		(d) the cut solids, and (e) the eroded solids. 
		}
		\label{anisotropic}
\end{figure}

\subsection{Adversarial Learning} 
Let us focus on how to train this framework now. Like other GANs, this framework takes adversarial learning, promoting the STG synthesizing more realistic solid textures to fool STDs.
When the adversarial learning is achieved, the STG can synthesize such realistic solid textures that the STDs cannot distinguish cross-sections of solid texture from the given exemplar.

It is worth noting that all discriminators at different scales are trained once in a single iteration, whereas the generator is trained only once at a randomly chosen scale, ensuring the learning is balanced.

This framework adopts the Wasserstein GAN with gradient penalty \cite{WGAN-GP} loss during optimization to improve the stability of training. When the generator is optimized, the loss is minimized using equation:
\begin{equation} 
 \mathcal{L}_G= -\mathbb{E}[D(u_n)],
 \label{eq1}
\end{equation}
where $u_n$ is a 2D slice taken at random from the 3D solid generated by STG. 
Meanwhile, each discriminator is optimized by minimizing its loss function, denoted by equation:
 \begin{equation}\begin{split}\label{eq2}
 \mathcal{L}_{D_n}&=\mathbb{E}[D(u_n)]-\mathbb{E}[D(x_n)]\\&+\lambda\mathbb{E}[(\parallel\nabla_{r_n} D(r_n)\parallel_2-1)^{2}],
 \end{split}
\end{equation}
where $x_n$ is a random cropped patch from the exemplar, and $r_n$ is a uniformly sampled data point between $u_n$ and $x_n$.

\subsection{3D Solid Texture Generator}
We adopt a fully convolutional network structure to carry out a solid texture generator, utilizing spatial locality by enforcing a local connectivity pattern between neurons of adjacent layers. 
The locality of pixel dependencies in fully convolutional operation can help STG to enforce the \emph{local Markov property} for generated solid textures.

In the STG, the trick of \emph{multi-scale inputs} \cite{20CNN,texturenet} is adopted, enabling the generator to learn textural details at different scales and capture the \emph{multiscality} of the texture.
In addition, the \emph{multi-scale inputs} influences the solid textural appearance at each scale to increase the diversity of texture and improving the stability of the adversarial learning.
Thus, at scale $n$, the input noise group $z_n$ for $G$ contains $K$ 3D noises $\left\{ z_{n, 1},\dots,z_{n, K}\right\}$ with different size.
\begin{equation}
v_n = G({\left\{ z_{n, 1},\dots,z_{n, K}\right\}}).
\end{equation}

\subsection{2D Slice Texture Discriminators}
\label{STDs}
The slice texture discriminator plays a critical role in guiding STG to produce realistic solid textures whose cross-section is largely indistinguishable from the given exemplar in the slicing direction. However, the scale of salient features could differ from each other in different directions. Furthermore, a single cross-section image may exhibit different spatial coherence at various scales. 
Although STG could generate textures with multiple resolutions, the \emph{multiscality} of texture requires an STS models to recognize features at multiple scales. Nevertheless, considering the complexity of textures, it is hard to train a discriminator to assemble multi-scale knowledge into a single model for differentiating multi-scale cross-sections.

Inspired by SinGAN \cite{SINGAN},  a set of slice texture discriminators are used to differentiate fake slices from given 2D exemplar at muti-scales.
In particular, unlike SinGAN's training mode from coarse-scale to fine-scale, STS-GAN trains the model on multiple scales simultaneously. Thus, the possible error accumulation from scale-by-scale training can be avoided.
The STDs consist of $N$ discriminators sharing the same fully convolutional structure but operating at different image scales. 
At the $n$ scale, $D_n$ takes the random slicing strategy to slice 2D cross-sections from corresponding synthesized 3D texture as ``fake" texture. In order to improve the diversity of real samples, we randomly crop textural patches at multiple predefined sizes from the exemplar and then resize these patches to the same resolution, providing the STDs with multi-scale ``real" textures.

\begin{table*}[t]
  \centering
  \scalebox{0.75}{
\begin{tabular}{cc|cc|ccc|ccc|ccc|ccc}
\hline
\multicolumn{2}{c|}{ExemplarA} & \multicolumn{2}{c|}{ExemplarB} & \multicolumn{3}{c|}{ExemplarC} & \multicolumn{3}{c|}{ExemplarD} & \multicolumn{3}{c|}{ExemplarE} & \multicolumn{3}{c}{ExemplarF} \\ \hline
Ours                & Chen      & Ours                & Chen      & {Ours}           & GramGAN & CNN   & {Ours}       & {GramGAN}       & CNN            & {Ours}           & {GramGAN} & CNN   & {Ours}  & {GramGAN} & CNN           \\ \hline
{0.559}     & 0.564     & {0.337}     & 0.368     & {1.87}  & 1.91  & 1.96  & 2.11  & {2.29}  & 2.52  & {2.76}  & 2.65  & 2.92  & 2.14  & 2.06  & {2.10} \\ \hline
\end{tabular}}
\caption{Comparison of FDMSE (×$10^{-4}$) on different exemplars. CNN means  \protect\cite{20CNN}. We calculate the FDMSE between the sections of the generated solids and the given exemplars. 2D exemplars are shown in Figure \ref{duibi1} and Figure \ref{duibi2}.}
\label{tab2}
\end{table*}

\begin{figure}[t]
 \centering
		\includegraphics[width=0.95\columnwidth]{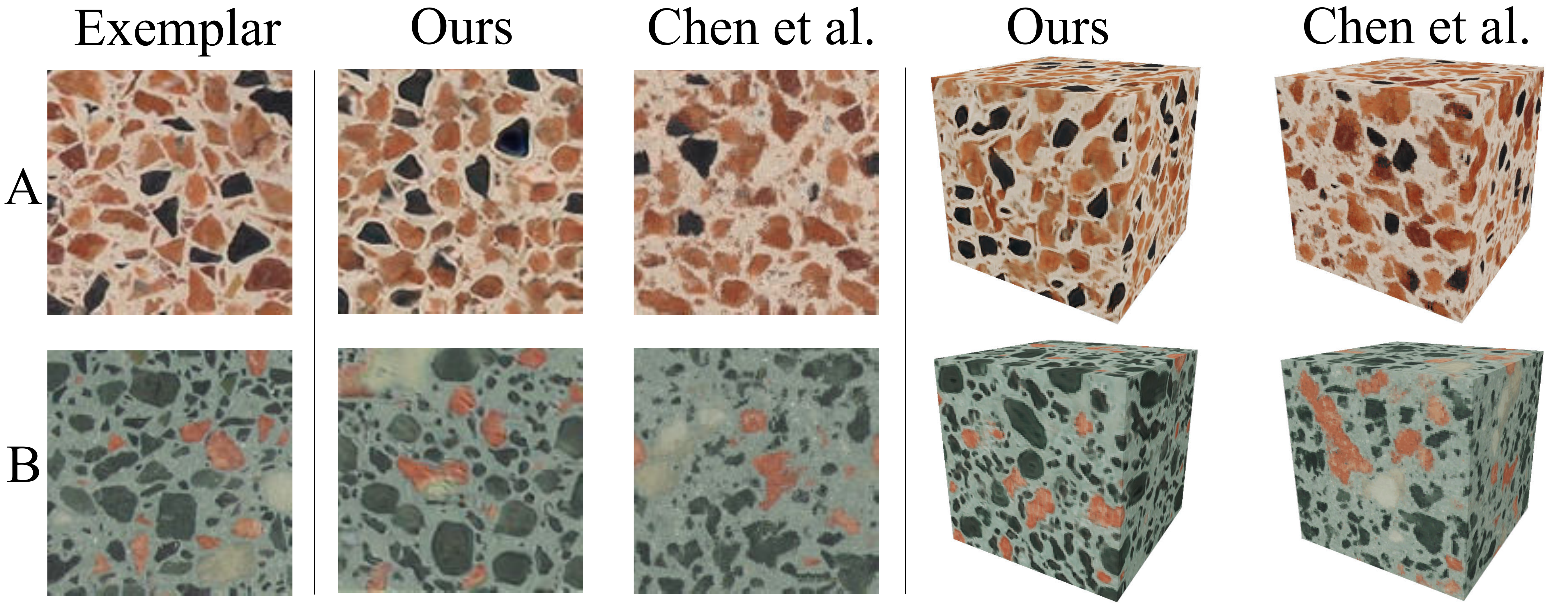}
		\caption{Comparison with the traditional non-neural method \protect\cite{Chen2010}. The synthetic slices are shown in the middle. The generated solids are shown on the right.}
		\label{duibi1}
\end{figure}

\section{Experiment}
\label{experiment}
This section examines the efficacy of our hierarchical architecture's multi-scale and orthogonal slicing components.
Isotropic and anisotropic experiments demonstrate the performance of our method.
Furthermore, we compare our method to the state-of-the-art STS methods, particularly those based on neural networks.

This framework runs on PyTorch and is optimized using the Adam optimizer \cite{kingma2014adam}. The learning rates of STG and STD are usually set to 0.0005 and 0.0003, respectively. The implementation uses Nvidia GeForce TITAN RTX GPU for acceleration.

\subsection{Ablation Experiments}
\textbf{Slicing Direction}
In the training process, to reduce computational overhead, we slice cross-sections in three orthogonal directions in STS-GAN. For comparison, we also slice additional 45-degree-angle cross-sections into the fake patch set to evaluate the sufficiency of the orthogonal slicing strategy.

Figure \ref{45du} shows the 3D textures obtained by the orthogonal slicing strategy and the 45-degree-added slicing strategy. There is no significant difference between these two slicing strategies, demonstrating that slicing along the orthogonal plane is sufficient to create high-fidelity solid textures.

\begin{figure*}[t]
	\centering
		\includegraphics[width=0.95\textwidth]{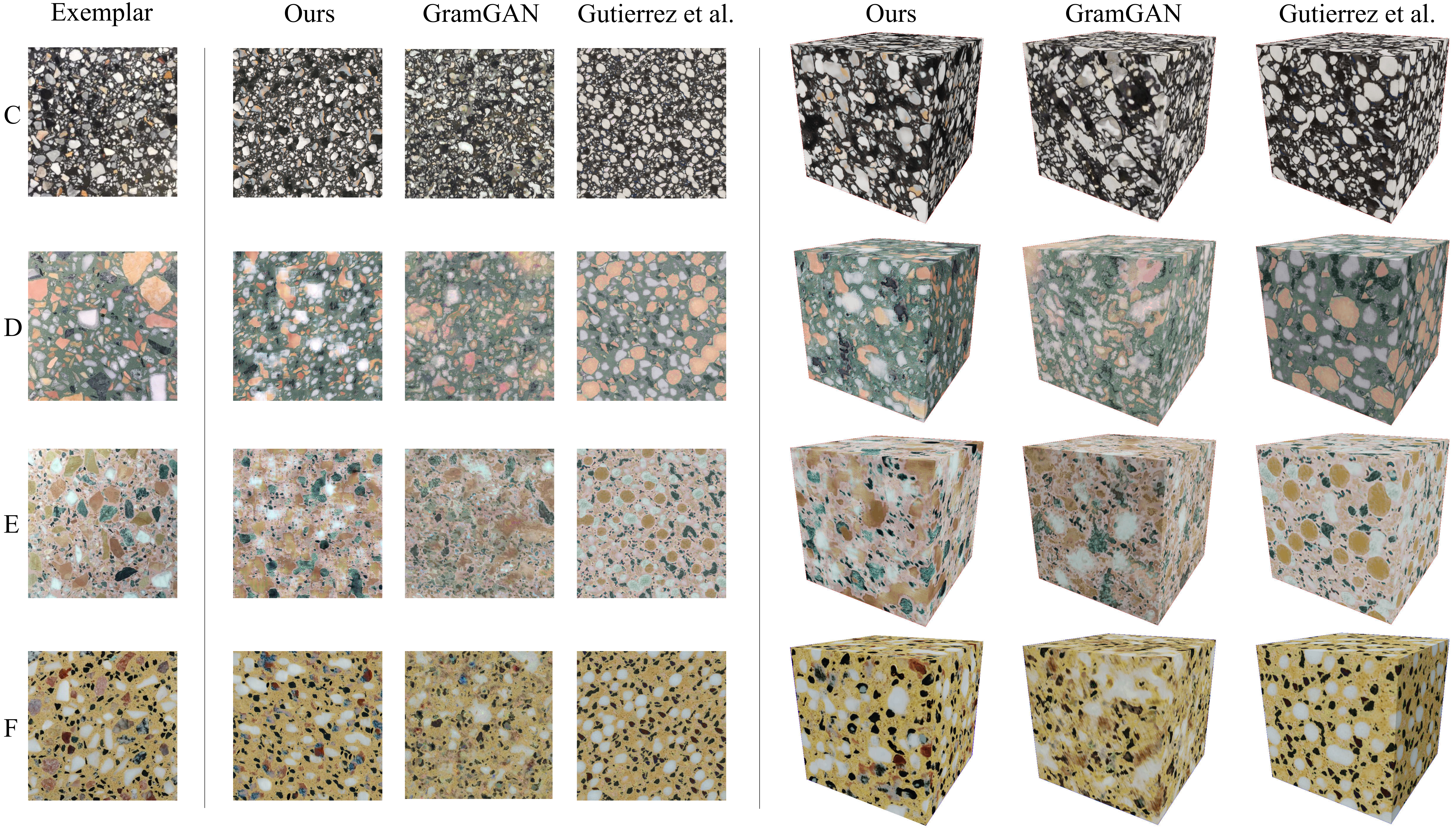}
		\caption{The comparison of STS-GAN's outcomes with those of neural networks-based methods. The generated solids are shown on the right. The slices are shown in the middle.}
		\label{duibi2}
\end{figure*}

\subsection{Parameter Analysis}
\textbf{Multi-scale Learning}
As previously stated, STS-GAN learns textural information on multiple scales. 
To confirm the efficacy of the multi-scale learning strategy, we explore its effects by varying the number of learning scales, and analyze model's  performance with various parameters (i.e., $N$).

Figure \ref{amd-1} shows that as the number of learning scales increases, the model provides clearer solids, and the overall structure and the local details gradually resemble the given exemplar.
Experiment proves that a model with multiple learning scales can capture the multiscality of textures and generate realistic solid textures at a sufficient learning scales (usually, $N=5$).

\begin{table}[t]
  
  \centering
  \scalebox{0.75}{
  \begin{tabular}{@{}c|ll@{}}
  \hline
  \multicolumn{1}{l|}{}                                                          & Method & Rank from Study \\ \hline
  \multirow{2}{*}{\begin{tabular}[c]{@{}c@{}}Non-neural \\ Networks\end{tabular}} & {Ours}   &   ${1.14}\ \pm{0.35}$             \\ 
 & Chen and Wang    &  $1.86 \ \pm 0.35$              \\ \hline
  \multirow{3}{*}{\begin{tabular}[c]{@{}c@{}}Neural \\ Networks\end{tabular}}     & {Ours}   &  ${1.47}\ \pm{0.74}$     \\  
 & GramGAN    &  $2.00\ \pm 0.68 $              \\ 
  & Gutierrez et al.     &      $2.53 \ \pm 0.66 $          \\ \hline
  \end{tabular}}
  \caption{User ranking for the similarity of the textures generated by the different methods to the exemplar. We report $\mu \pm \sigma$ (mean and standard deviation)  for user study ranking (lower is better).}
  \label{userstudy}
  \end{table}

\subsection{Dependency of Direction}
\textbf{Isotropic Exemplar}
In experiments, the STS-GAN is evaluated with isotropic exemplars. Figure \ref{isotropic} depicts the generated solid texture and several slices of solid.
It is apparent that the created 3D solid closely resembles the given 2D exemplar and the textural visual characteristics of the slices from the 3D solid at different angles are similar to the exemplar.

Furthermore, the generated solid is used in surface texture mapping. Based on the spatial coordinate information, the synthesized solid distributes color to the surface pixels of the 3D mesh model. The outcome exhibits similar visual qualities to the exemplar, as shown in Figure \ref{isotropic}(g).

These experiments show that STS-GAN can learn texture distribution of the given exemplar and synthesize high-fidelity solid textures.

\textbf{Anisotropic Exemplar}
Since the STS-GAN learns the texture distribution from a 2D exemplar, it can also generate solid textures with anisotropic properties by capturing different exemplars. In experiments, the STS-GAN learns anisotropic exemplars in multiple orthogonal orientations based on different configurations.

As shown in Figure \ref{anisotropic}, the generated solid textures maintain the same texture properties in each orthogonal direction as the corresponding exemplar, and we present the synthetic solids differently. 
Experiments suggest that STS-GAN is able to learn anisotropic texture and extend it to 3D solids.

\subsection{Performance Comparison}
In this section, the performance of the STS-GAN is compared against three state-of-the-art methods, including a non-neural method \cite{Chen2010} and two neural networks-based methods \cite{20CNN,GramGAN}.

\textbf{Qualitative Evaluation}
Figure \ref{duibi1} compares our method with Chen et al.'s method. Although Chen et al.'s approach produces solid textures similar to exemplars, they still have failed attempts with variances between the generated solid and the exemplar. 
It can be observed that the solid texture produced by STS-GAN is more similar to the exemplar. Furthermore, the STS-GAN provides clearer borders and textures consistent in color, shape, and distribution with the exemplars.
Due to the powerful learning ability of neural networks, STS-GAN has a more remarkable ability to learn textural properties than non-neural methods. Thus, our method can capture complex textures and generate realistic 3D textures.

% \color{black}
Figure \ref{duibi2} exhibits the comparison between STS-GAN and two neural networks-based methods. It can be observed that the solid textures and slices generated by STS-GAN are more visually similar to the exemplar.  
In most cases, Gutierrez et al.' method focuses solely on the generalized textural styles.
For textures with diverse pattern styles, their approach fails to adapt to the \emph{diversity} of textures.
In contrast, STS-GAN can learn arbitrary diverse texture patterns and extend them to solid textures benefiting from the ability of GAN to approximate arbitrary distributions.
As shown in Figure \ref{duibi2}, solid textures generated by GramGAN are blurred, and the color and shape of textures differ from the exemplar. 
These situations are due to the difficulty of GramGAN in learning the \emph{local Markov property} and \emph{multiscality} of textures.
In STS-GAN, the CNNs-based network structure ensures that STS-GAN can capture the \emph{local Markov property} of textures. At the same time, the multi-scale strategy helps STS-GAN to learn textural \emph{multiscality}.
Thus, our method produces high-fidelity solid textures compared to the other two methods.

\textbf{Frequency Domain Mean Square Error}
In this section, we quantitatively compare STS-GAN with other competitors. 
In the experiments, we use the frequency domain mean square error (FDMSE) as metric to evaluate synthesis effect. The frequency domain of an image contains information of the texture \cite{2019FD,FDtexture}.
Specifically, during the computational process, we first calculate the magnitude based on the image frequency domain, and then proceed with the calculation of mean squared error.
Table \ref{tab2} shows the performance on the this metric. 
It can be observed that the neural network methods significantly outperform the traditional ones on FDMSE.
Particularly, STS-GAN and GramGAN are superior to CNN in multiple exemplars, as they adopt GAN, enabling capturing diverse textural information, generating visually similar solid textures.

Compared to image numerical metrics, the user study provides a closer approximation to the visual quality.
Thus, to evaluate different methods more intuitively and effectively, we conducted a user study experiment as described below.

\textbf{User Study}\footnote{All volunteers have signed an informed consent form, guaranteeing to make independent and objective choices.}
We also conducted a single-blind formal user study to compare the visual effect between approaches. A group of 26 volunteers was given texture exemplars and their corresponding slices from synthesized solid textures by different competitors. They were asked to rank the slices based on their similarity to the corresponding reference exemplar. Apart from their corresponding exemplars, the questionnaire contains 20 groups of slices from non-neural network synthesizers and 20 groups from neural network ones. Table \ref{userstudy} exhibits the average ranking for each approach. 
It can be observed that the average ranking of our method is significantly higher than the other methods, and users are more accepting of our results. 
Experimental results prove our method can generate more realistic solid textures.

\section{Conclusion}
\label{Conclusion}
  This research proposes a novel approach to synthesize solid texture, named as STS-GAN, which learns arbitrary 3D solid texture from few 2D exemplars.
  It can successfully capture textural \emph{local Markov property}, \emph{multiscality}, and \emph{diversity}, synthesizing high-fidelity solid textures. 
  In experiments, STS-GAN generates more realistic solid textures than the other state-of-the-art methods.

  There are, however, still some limitations to this method. The high computational cost of STS-GAN learning process is a significant burden. In the future, the simplification method should be further studied to accelerate learning.
  Furthermore, the generating process must be hastened to meet the efficiency requirements in real-world applications.
  Finally, there is a future research direction in exploring the relationship between texture and the human visual perception, as well as establishing precise numerical metrics for describing textures.

\color{black}

\section*{Acknowledgments}
This work was supported by National Natural Science Foundation of China under Grant No. 61872419, No. 62072213, No. 61873324, No. 61903156. Shandong Provincial Natural Science Foundation No. ZR2022JQ30, No. ZR2022ZD01, No. ZR2020KF006. Taishan Scholars Program of Shandong Province, China, under Grant No. tsqn201812077. “New 20 Rules for University” Program of Jinan City under Grant No. 2021GXRC077.
Project of Talent Introduction and Multiplication of Jinan City.

\bibliographystyle{named}
\bibliography{ijcai23.bib}

\end{document}